\documentclass[10pt,twocolumn,letterpaper]{article}

\usepackage[pagenumbers]{wacv} 

\usepackage{graphicx}
\usepackage{amsmath}
\usepackage{amssymb}
\usepackage{booktabs}
\usepackage{algorithm}
\usepackage{algorithmic}
\usepackage{multirow, multicol}

\usepackage[pagebackref,breaklinks,colorlinks]{hyperref}

\usepackage[capitalize]{cleveref}
\crefname{section}{Sec.}{Secs.}
\Crefname{section}{Section}{Sections}
\Crefname{table}{Table}{Tables}
\crefname{table}{Tab.}{Tabs.}

\def\wacvPaperID{248} 
\def\confName{WACV}
\def\confYear{2025}

\begin{document}

\title{LIPIDS: Learning-based Illumination Planning In Discretized (Light) Space for Photometric Stereo}

\author{Ashish Tiwari, Mihir Sutariya, and Shanmuganathan Raman\\
CVIG Lab, IIT Gandhinagar\\
{\tt\small \{ashish.tiwari, 
sutariya.mihirkumar, shanmuga\}@iitgn.ac.in}
}
\maketitle

\begin{abstract}

Photometric stereo is a powerful method for obtaining per-pixel surface normals from differently illuminated images of an object. While several methods address photometric stereo with different image (or light) counts ranging from one to two to a hundred, very few focus on learning optimal lighting configuration. Finding an optimal configuration is challenging due to the vast number of possible lighting directions. Moreover, exhaustively sampling all possibilities is impractical due to time and resource constraints. Photometric stereo methods have demonstrated promising performance on existing datasets, which feature limited light directions sparsely sampled from the light space. Therefore, can we optimally utilize these datasets for illumination planning? In this work, we introduce LIPIDS - \textbf{L}earning-based \textbf{I}llumination \textbf{P}lanning \textbf{I}n \textbf{D}iscretized light \textbf{S}pace to achieve minimal and optimal lighting configurations for photometric stereo under arbitrary light distribution. We propose a Light Sampling Network (LSNet) that optimizes lighting direction for a fixed number of lights by minimizing the normal loss through a normal regression network. The learned light configurations can directly estimate surface normals during inference, even using an off-the-shelf photometric stereo method. Extensive qualitative and quantitative analyses on synthetic and real-world datasets show that photometric stereo under learned lighting configurations through LIPIDS either surpasses or is nearly comparable to existing illumination planning methods across different photometric stereo backbones. 
\end{abstract}

\begin{figure}[!h]
\centering
  \includegraphics[width=0.9\linewidth]{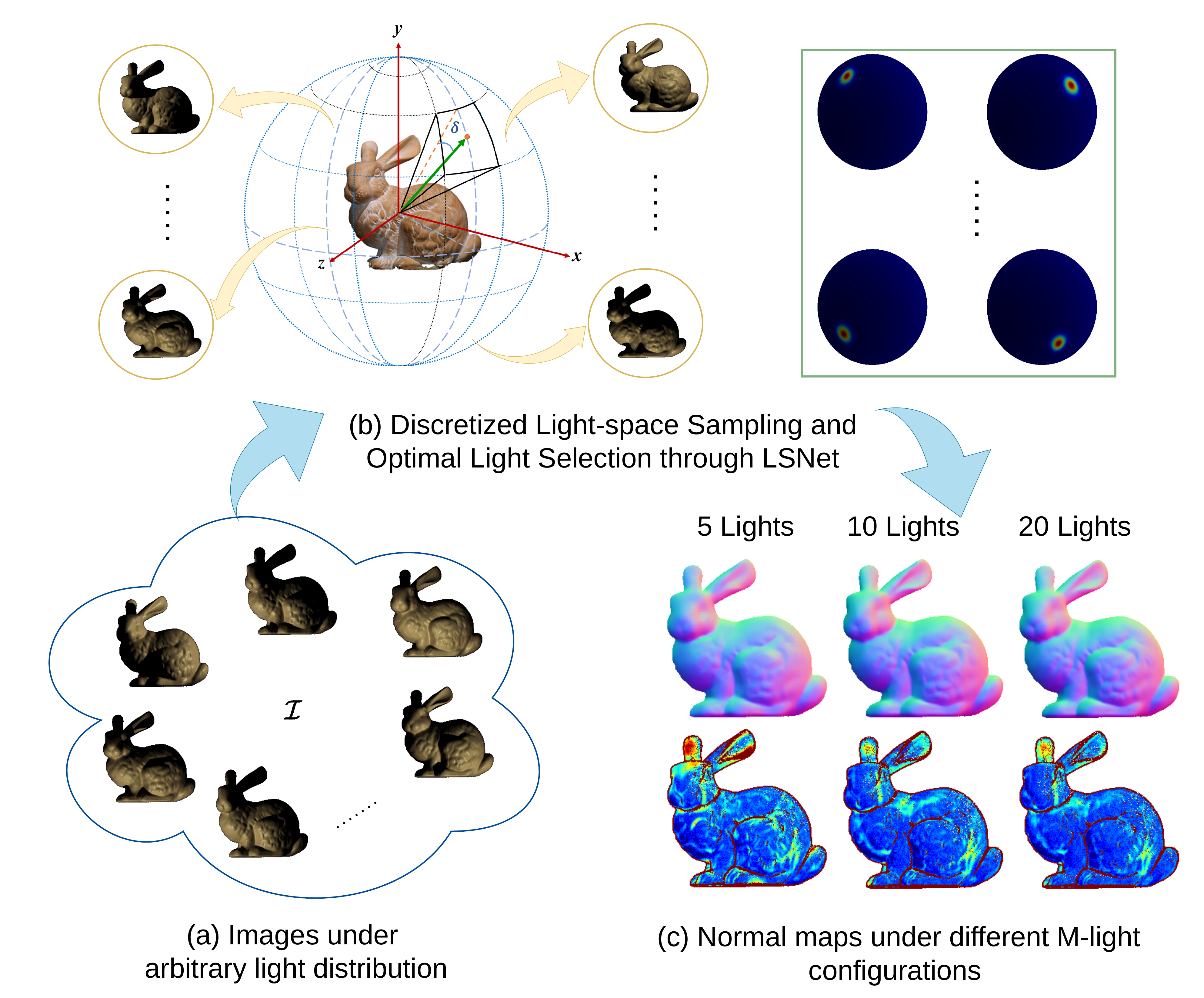}
  \caption{Schematic diagram of the proposed method for illumination planning in photometric stereo. (a) Given a set of images under an arbitrary lighting distribution, we first discretize the light space into $K$ bins spanning the upper hemisphere, (b) followed by light assignment to an appropriate bin. For a desired $M < K$, we use Light Sampling Network (LSNet) to select an optimal $M$-light configuration such that (c) the normals estimated using this set of $M$ lights have the least mean angular error.}
  \label{fig:teaser}
  \vspace{-1mm}
\end{figure}

\section {Introduction} \label{sec:intro}

Surface normals are one of the many ways to characterize a surface. They are widely used across different fields, such as in computer graphics for extensive visualizations and sophisticated renderings, in industries for material inspection, and even in robotics for scene understanding. Photometric stereo is one of the most popular methods in computer vision to estimate the per-pixel surface normal of an object by analyzing its images under multiple different lightings. Woodhman first introduced photometric stereo in 1980 \cite{woodham1980photometric}, assuming a pure Lambertian surface and three-light (non-co-planar) setup. However, such consideration is highly restrictive since objects in the real world exhibit complex surface reflectance, and their appearance is heavily influenced by global illumination effects such as shadows and inter-reflections. 

Several photometric stereo methods \cite{tiwari2022lerps, tiwari2022deepps2, ikehata2018cnn,  zheng2019spline, santo2017deep, ikehata2023scalable,chen2018ps, chen2019self, lichy2021shape} have shown improved normal estimation accuracy by considering images under more than three light directions. Interestingly, the spectrum of the required number of lights is very wide, ranging from as little as one (Shape from Shading), two, three, and ten lights to close to a hundred lights. While methods have been designed to perform under an arbitrary number of lights, no single number is available that applies to a large set of objects in general. While several physical constraints and assumptions are required when dealing with fewer images (such as a co-located camera-light setup) \cite{lichy2021shape}, acquiring more images often demands carefully orchestrated setups involving controlled lighting environments and precise calibration procedures \cite{debevec2000acquiring}. Due to several practical constraints such as time, cost, and equipment limitations, it is often infeasible to exhaustively sample the entire space of possible lighting configurations. As a result, sometimes, the acquired dataset may not sufficiently cover all relevant lighting variations, leading to incomplete or inaccurate surface reconstructions.
Furthermore, images captured under certain lighting conditions may not yield optimal results even after extensive data collection due to underlying redundancies. This implies that an alternative set of lighting conditions, possibly fewer in number, could produce similar, if not better, outcomes. To account for these observations and address the aforementioned challenges, it is important to design a scheme to select the optimal lighting configuration well before the data acquisition and strike a balance between accuracy and preferably a minimum number of light directions. It also proves useful for scaling photometric stereo to more general environment settings under minimal optimal lighting. 

While exhaustive sampling of light space is impractical, the existing photometric stereo baselines have validated their performance on standard photometric stereo datasets \cite{alldrin2008photometric, debevec2006relighting, chen2018ps, shi2016benchmark, wang2023diligent, ren2022diligent102, Guo_Ren_Wang_2024_CVPR}. However, these datasets are acquired with relatively sparse light space sampling. Now the question is - \textit{can we use the existing datasets to find a global optimal lighting configuration that could apply to a large variety of objects?} Surprisingly, despite the heavily discussed and recognized importance, only a handful of methods have addressed illumination planning for photometric stereo \cite{drbohlav2005optimal,tanikawa2022online, gardi2022optimal,chan2023releaps}. Some existing methods have either relied on handcrafted priors or simplifying assumptions on reflectance models \cite{drbohlav2005optimal,tanikawa2022online, gardi2022optimal}, while others \cite{chan2023releaps} have used reinforcement learning for illumination planning. 

\textbf{Contributions.} To address generalized photometric stereo in the real world, it is important that the illumination planning considers its dependence on the shape, reflectance, and global illumination effects and can be deployed with any photometric stereo pipeline.

(a) In this work, we propose a simple and effective Light Sampling network (LSNet) that learns to select the optimal lighting configuration (i.e., position in the light space) for a given number of lights in the discretized light space for illumination planning in an offline manner.

(b) We observe that the proposed illumination planning framework (LIPIDS) either surpasses (or nearly matches) the performance of the existing illumination planning methods. Moreover, the proposed method can easily be integrated with the existing photometric stereo benchmarks. 

(c) Through an extensive evaluation of the synthetic and real datasets, we demonstrate the efficacy of the LIPIDS (via LSNet) in obtaining universal optimal lighting configurations that apply to any lighting distribution and allow faster inference. 

\begin{figure*}[t]
    \centering
    \includegraphics[width=\linewidth]{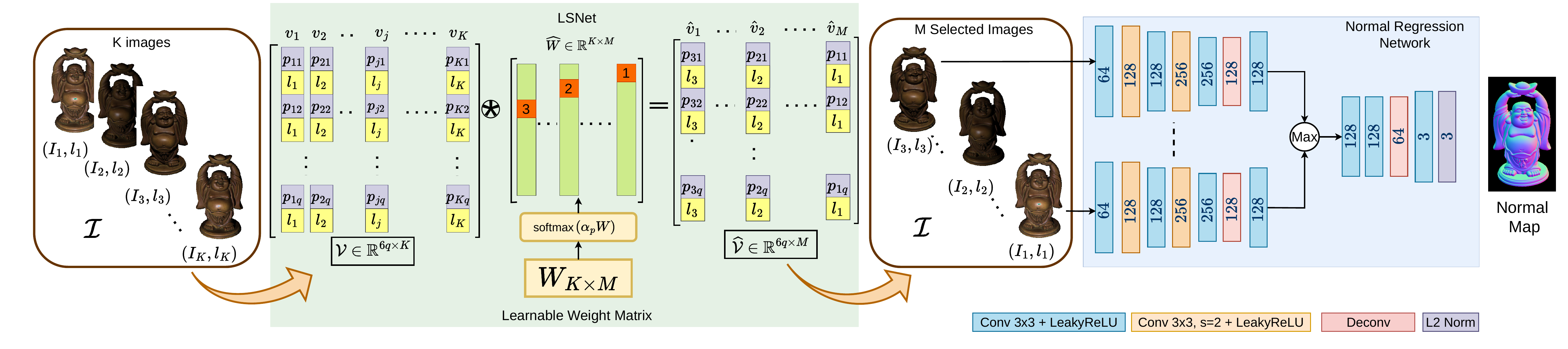}
    \caption{Architecture of the LIPIDS framework with Light Sampling Network (LSNet) and Normal Regression Network (NRNet) }
    \label{fig:1}
\end{figure*}

\section{Related Work} \label{sec:related_work}

In this section, we discuss the existing photometric stereo methods that use different numbers of images to estimate surface normals and describe the existing illumination planning methods.

\textbf{Learning-based Photometric Stereo.} Although the classical photometric stereo requires a minimum of three images captured under non-coplanar light directions, it applies only to Lambertian assumption \cite{woodham1980photometric}. This method requires a greater number of images under varying lights to handle general reflectance and global illumination effects. Some learning-based methods have adopted different training strategies to address photometric stereo using one \cite{tiwari2022lerps} or two \cite{tiwari2022deepps2} images during inference. However, they require more images during training. Photometric stereo has been approached through per-pixel \cite{ikehata2018cnn} and all-pixel methods \cite{chen2018ps}. Most of these methods have shown better performance with around $50 \sim 100$ images by designing different network architectures using convolutional neural networks (CNN) \cite{santo2017deep, chen2019self, ikehata2023scalable, logothetis2021px, ikehata2018cnn}, transformers \cite{ikehata2022ps} or graph convolution networks (GCNs) \cite{yao2020gps} to handle general BRDFs. Interestingly, Zhang \emph{et al.} \cite{zheng2019spline}, and Li \emph{et al.} \cite{li2019learning} designed different networks for photometric stereo under sparse light distribution. However, they first use sparse lighting to extrapolate to dense observation maps and then use them to compute surface normals. Since the literature on photometric stereo is very wide, we refer the readers to \cite{ackermann2012photometric, ju2024deep} for more detailed discussions. 

\textbf{Illumination Planning.} Finding optimal lighting configurations has been approached offline and online. While offline illumination planning methods offer a universal configuration that applies to various objects, online strategies tailor lighting configurations to specific objects with more capture time and high sensitivity to noise. Illumination planning under an offline setting was first addressed by Drbohlav \emph{et al.} \cite{drbohlav2005optimal} for Lambertian photometric stereo in the presence of camera noise. They show that for three lights, any triplet of orthogonal light directions is optimal. Later, Tanikawa \emph{et al.} \cite{tanikawa2022online} introduced an online method to find optimal light directions iteratively by minimizing the effects of shadow on surface normal accuracy. Further, Gardi \emph{et al.} combined calibrated photometric stereo with optimization-based parameter estimation and optimal experimental design. Iwaguchi \emph{et al.} \cite{iwaguchi2023surface} proposed an offline approach for non-Lambertian surfaces applicable only for near-field photometric stereo. Recently, Chan \emph{et al.} \cite{chan2023releaps} proposed an online method based on reinforcement learning, considering an iterative optimization for generalized photometric stereo by integrating reward functions in learning. However, their per-object optimization is time-consuming during inference and requires a sophisticated imaging setup. Unlike these works, we propose a learning-based offline illumination planning strategy for generalized photometric stereo based on light space discretization that can be applied to any lighting distribution and allow faster inference.

\section{Method} \label{sec:method}

This section describes the image formation model for photometric stereo and discusses illumination planning by sampling in the discretized light space.

\subsection{Illumination Planning in Photometric Stereo}
Let us consider an anisotropic non-Lambertian surface $f$ characterized by the Bidirectional Reflectance Distribution Function (BRDF) $\rho$ imaged by an orthographic camera with a linear radiometric response. For a surface point $\boldsymbol{x} \in \mathbb{R}^{3}$ mapped onto pixel $\boldsymbol{p} \in \mathbb{R}^{2}$ in the image, let $\boldsymbol{n} \subset \mathcal{S}^{2} \in \mathbb{R}^{3}$ be the surface normal. Suppose the point is illuminated by the light source in the direction $\boldsymbol{l} \in \mathbb{R}^{3}$ and viewed from the direction $\boldsymbol{v} \in \mathbb{R}^{3}$. The resulting image intensity (normalized by the light intensity) is given by Equation \ref{eq:1}.
\begin{equation}
    \boldsymbol{I}(\boldsymbol{p}) = \rho(\boldsymbol{n}, \boldsymbol{l}, \boldsymbol{v}) \psi(\boldsymbol{p})[\boldsymbol{n}(\boldsymbol{p})^{T}\boldsymbol{l}] + \epsilon
    \label{eq:1}
\end{equation}
While $\epsilon$ accommodates inter-reflections and noise, $\psi(\boldsymbol{p})$ represents the underlying cast and attached shadows. $\psi(\boldsymbol{p})= 0$, if $\boldsymbol{p}$ is shadowed, otherwise, it equals $1$.

Generalized photometric stereo considers a total of $K \geq 3$ images under different lightings to obtain the surface normal map $\boldsymbol{N} \in \mathbb{R}^{H \times W \times 3}$ for a general non-Lambertian surface with global illumination effects and noise. Let $\mathcal{I} = \{\boldsymbol{I_{1}},\boldsymbol{I_{2}}, ...,  \boldsymbol{I_{K}}\}; \boldsymbol{I}_{k} \in \mathbb{R}^{H \times W \times 3}$ be the set of images under $K$ different lightings $\boldsymbol{L} = \{l_{1}, l_{2}, ..., l_{K}\} \in \mathbb{R}^{K \times 3}$. Then, obtaining surface normal from photometric stereo (\textsf{PS}) is equivalent to solving Equation \ref{eq:2}.
\begin{equation}
    \boldsymbol{N} = \textsf{PS}(\mathcal{I}, \boldsymbol{L})
    \label{eq:2}
\end{equation}

Illumination planning attempts to find an optimal light configuration of $M$ lights out of a given $K$-light distribution such that $\boldsymbol{L}_{opt} = \boldsymbol{L}_{M} \subseteq \boldsymbol{L}_{K}$. The optimal light configuration $\boldsymbol{L}_{opt}$ is obtained by minimizing the resulting mean angular error with respect to the ground truth normal map $\boldsymbol{N}_{gt}$, such that the Equation \ref{eq:3} holds.

\begin{equation}
    \boldsymbol{L}_{opt} = \underset{\boldsymbol{L}_{M}}{\text{argmin}}  \frac{1}{HW} [1 - (\textsf{PS}(\mathcal{I}_{M}, \boldsymbol{L}_{M}) \cdot \boldsymbol{N}_{gt})]
    \label{eq:3}
\end{equation}

The schematic of the proposed framework is described in Figure \ref{fig:teaser}.

\subsection{Proposed Method: The LIPIDS Framework}

Several learning-based methods \cite{zheng2019spline, li2019learning} have explored photometric stereo using a sparse set of images and lights from specific directions. However, no prior information is available regarding the optimal lighting configuration. One choice for learning the optimal lighting directions is to directly regress the lighting position using a learning-based network. However, changes in light direction can result in complex changes in the scene that are difficult to model and often are not differentiable with respect to the lighting, for example, tracking changes in shadows due to changing light source position. Moreover, identifying the optimal light configuration involves navigating a vast search space, typically the upper hemisphere of the object, which is practical for capturing distinct shading variations without significant occlusions or shadows. While this involves a sophisticated acquisition setup, the upper bound on the number of images is not fixed. While one solution could be to use a differentiable image renderer to synthetically generate images under different lightings, the physical correctness of the rendered images is often compromised by the underlying assumptions of the renderer. 

\textbf{\textit{Do we need the exact lighting directions?}} Under distant lighting, specifying a region in the light space rather than the exact direction is found to perform well  \cite{chen2019self, tiwari2022deepps2}. A few recent studies \cite{chen2019self, tiwari2022deepps2} have shown that modelling lighting estimation as a classification task through light space discretization helps to reduce learning difficulties in the network and allows the downstream tasks (such as normal estimation) to tolerate errors in estimated lightings better. Moreover, this allows for an easier light source calibration for real data acquisition. The idea of discretization in the context of photometric stereo has also been exploited by Enomoto \emph{et. al.} \cite{enomoto2020photometric}. However, they discretize surface normals and BRDFs in contrast to discretizing light space.

\begin{figure}[h]
\centering
\begin{tabular}{cc}
       \includegraphics[width=\linewidth]{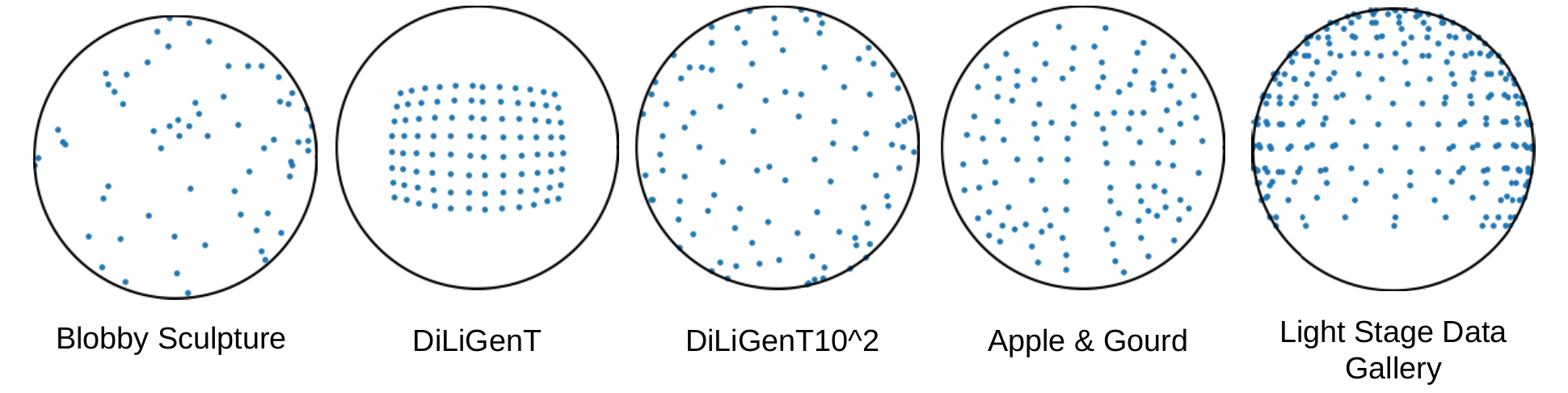} & \\
      \includegraphics[width=0.9\linewidth]{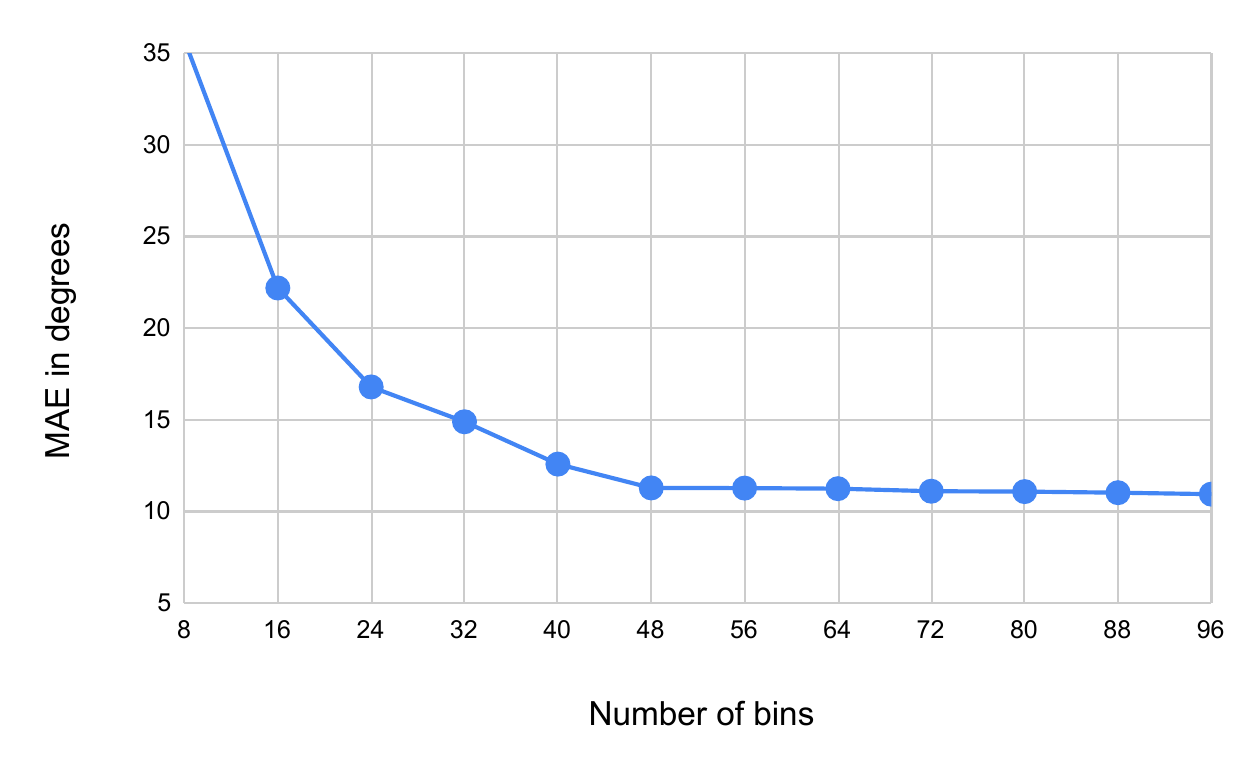}    
\end{tabular}
\caption{Top: Different lighting distributions across the publicly available photometric stereo datasets. Bottom: Variation of MAE with different numbers of light bins evaluated over the DiLiGenT dataset\cite{shi2016benchmark}. Note that for all values of $K$, we sample more along azimuth than elevation.}
\label{fig:light_dist}
\vspace{-2mm}
\end{figure}

Drawing insights from these observations, we propose to perform illumination planning in discretized light space. Moreover, instead of capturing the dataset under multiple different lightings by densely sampling the light space, we propose to optimally utilize the lighting distributions available in the Blobby \& Sculpture dataset \cite{chen2018ps} to learn a universal lighting configuration that generalizes to any lighting distribution available during the test time. The lighting distributions of the publicly available photometric stereo datasets are shown in Figure \ref{fig:light_dist}). Blobby \& Sculpture dataset \cite{chen2018ps} is a standard dataset consisting of synthetic images with complex normal distributions and materials from MERL dataset \cite{matusik2003data}. Most importantly, it contains images captured under near-uniform lighting distribution over the upper hemisphere. 


\begin{figure}[h]
    \centering
    \includegraphics[width=\linewidth]{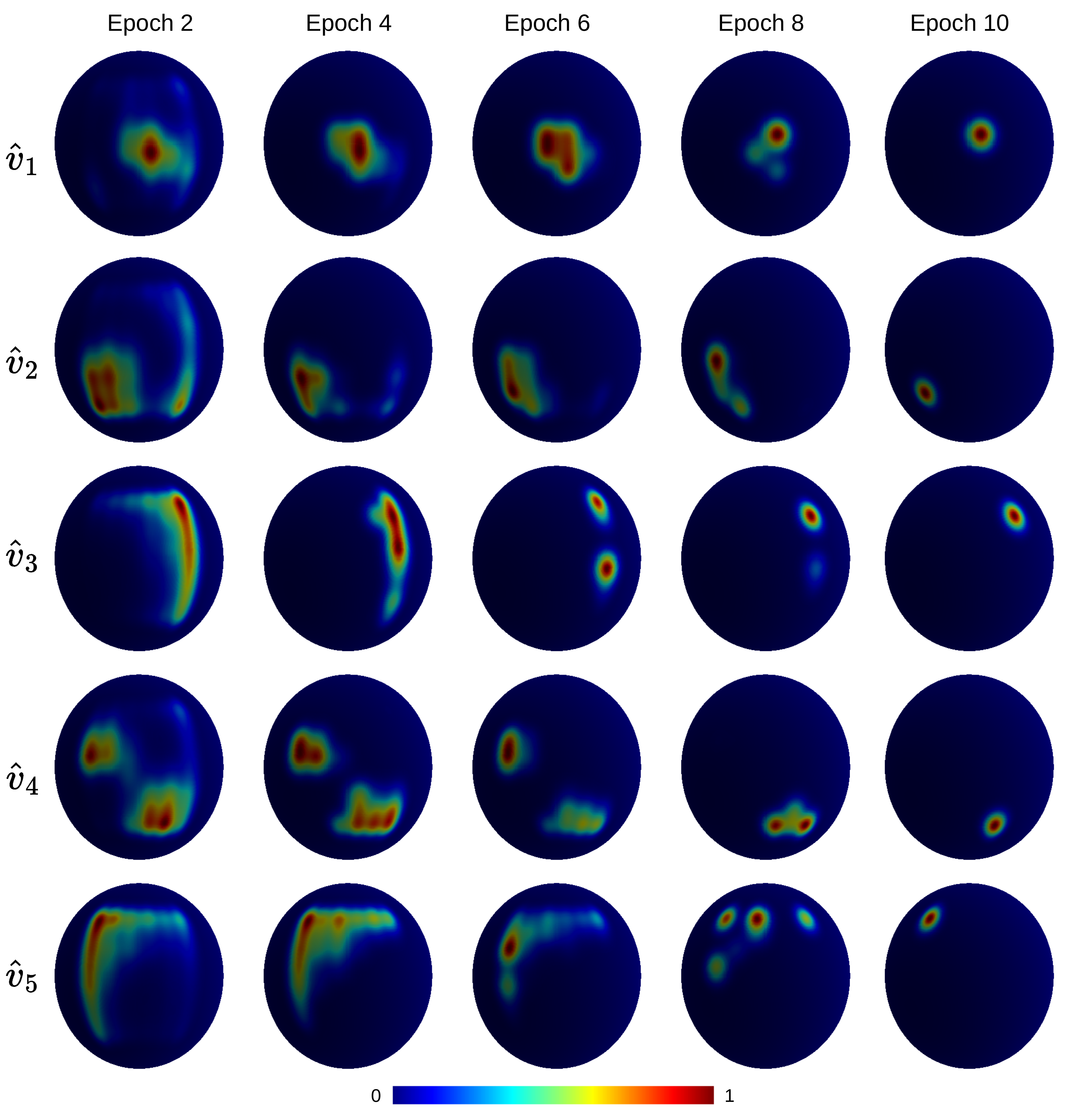}
    \caption{ Evolution of the optimal lighting configuration while training LSNet + NRNet over ten epochs. Each row represents the values in one column of $\widehat{W}$ matrix. The final converged light position represents the associated light bin in the discretized light space.}
    \label{fig:light_evol}
\end{figure}

\textbf{Light-space Discretization.} Given a set of images captured under an arbitrary lighting distribution, we first discretize the light space into $K=48$ light bins and assign the lights to the appropriate bins. The objective is to find $M$ out of $K$ light bins that minimize Equation \ref{eq:3}. While a larger $K$ value requires a larger number of images (possibly redundant images), a smaller $K$ might not span the upper hemisphere adequately. We analyzed the error curve over the DiLiGenT dataset for different numbers of bins under $10-$light configuration, as shown in Figure \ref{fig:light_dist}. The MAE is averaged across three PS backbones LS \cite{wu2011robust}, PS-FCN \cite{chen2018ps}, and CNN-PS \cite{ikehata2018cnn}. While we observed that the error curve starts to flatten for $K \geq 32$, the range from $[32,64]$ is reasonably applicable for light distributions across different datasets. Moreover, these bins provide good coverage of the upper hemisphere. With the Blobby \& Sculpture dataset having $64$ images per object, we chose $K=48$ to fall well within the expected range for our experiments.

The lighting directions in the upper hemisphere can be described by azimuth angle $\phi \in [0^{\circ}, 180^{\circ}]$ and elevation angle $\theta \in [-90^{\circ},90^{\circ}]$. We discretize the light direction space by dividing the azimuth into eight bins ($K_{\phi} = 8$) and the elevation into six bins ($K_{\theta} = 6$), resulting in a total of 48 bins ($K = K_{\phi} \times K_{\theta} = 48$). We observed that lighting variation is more pronounced along the azimuth direction than the elevation direction, leading us to choose uneven discretization. This approach also benefits from a reduced number of bins. Interestingly, upon dividing the space evenly with eight bins along both directions,  we obtained a similar mean absolute error (MAE) (see Figure \ref{fig:light_dist}). The setup bears a maximum angular deviation of $11.25^{\circ}$ and $15^{\circ}$ along the azimuth and elevation direction, respectively.

\textbf{Light Bin Assignment.} The simplest way to allocate light to a particular bin is to pick one with the smallest angular deviation with the given bin direction, which works well for uniform or near-uniform light space sampling. However, especially for non-uniform or biased light space sampling, such a strategy causes multiple bins to be assigned to the same light (collisions and reduced span of the light space) or one bin to have more than one light.  To ensure optimal assignment and avoid such ambiguities, we propose a simple light assignment strategy, described by Algorithm \ref{algo:1}.


\begin{algorithm}[t]
\caption{Light Bin Assignment}
\begin{algorithmic}[1]
    \STATE \textbf{// Initialize bin and light directions}
    \STATE \textsf{bins\_directions} $\gets$ list of bin directions
    \STATE \textsf{light\_directions} $\gets$ list of light directions
    
    \STATE \textbf{// Calculate angles between bin and light directions}
    \STATE \textsf{Angles} $\gets$ empty list
    \FOR{each bin direction $i$ in \textsf{bins\_directions}}
        \FOR{each light direction $j$ in \textsf{light\_directions}}
            \STATE \textsf{angle} $\gets$ \textsf{calculate\_angle}(\textsf{bins\_directions}$[i]$, \textsf{light\_directions}$[j]$)
            \STATE append (\textsf{angle}, $i$, $j$) to \textsf{Angles}
        \ENDFOR
    \ENDFOR
    
    \STATE \textbf{// Sort the angles in increasing order}
    \STATE sort \textsf{Angles} by \textsf{angle} in ascending order
    
    \STATE \textbf{// Assign lights to bins using a greedy algorithm}
    \STATE \textsf{bin\_light\_pairs} $\gets$ array of size \textsf{bins\_directions} initialized to -1
    \FOR{each (\textsf{angle}, \textsf{bin\_index}, \textsf{light\_index}) in \textsf{Angles}}
        \IF{\textsf{bin\_light\_pairs}$[$\textsf{bin\_index}$] == -1$}
            \STATE \textsf{bin\_light\_pairs}$[$\textsf{bin\_index}$] \gets$ \textsf{light\_index}
        \ENDIF
    \ENDFOR
    
    \STATE \textbf{Output}: \textsf{bin\_light\_pairs}, where each bin is assigned a light direction index.
\end{algorithmic}
\label{algo:1}
\end{algorithm}

\subsection{Light Sampling Network (LSNet)}
Consider a set of image light pairs $ \mathcal{S} = \{(\boldsymbol{I_{1}}, \boldsymbol{l}_{1}) ,(\boldsymbol{I_{2}}, \boldsymbol{l}_{2}) , ...  (\boldsymbol{I_{K}}, \boldsymbol{l}_{K})\}$ obtained after light space discretization and bin assignment. For an image $\boldsymbol{I}_{j}$, we append the lighting direction $\boldsymbol{l}_{j} \in \mathbb{R}^{3}$ to every $i^{th}$ pixel $\boldsymbol{p}_{ij} \in \mathbb{R}^{3}$ of $\boldsymbol{I}_{j}$ and create a vector $\boldsymbol{v}_{j} \in \mathbb{R}^{6q}$. Here, $q$ is the number of pixels per image channel. We stack the paired image-light information (encoded in vectors $\boldsymbol{v}_{j}$) as the columns of matrix $\mathcal{V} \in \mathbb{R}^{6q \times K}$. We can now define the selection of $M$ optimal lights as per Equation \ref{eq:4}.
\begin{equation}
    \widehat{\mathcal{V}} = \mathcal{V}\widehat{W}
    \label{eq:4}
\end{equation}
Here, $\widehat{W} \in \mathbb{R}^{K \times M}$ is a binary matrix, such that each column of $\widehat{W}$ has only a single entry as $1$ corresponding to the image (or light) that is selected from $\mathcal{V}$. Technically, as shown in Figure \ref{fig:1}, LSNet is a framework consisting of light selection through a learnable weight matrix $\mathcal{W}$ (initialized to all 1s before training) that selects a sparse set of light samples from $\mathcal{V}$. To ensure that the multiplication matrix is binary and has a single $1$ entry per column, inspired by \cite{chakrabarti2016learning, xu2018deep}, we apply the \textsf{softmax} operation to each column of $\mathcal{W}$, such that $\widehat{\mathcal{W}} = \texttt{softmax}(\alpha_{r}\mathcal{W})$. Here, $\alpha_{r}$ is a scalar that gradually increases from $1$ to a very large value during each epoch $r$ during training, following $\alpha_{r} = \beta{r}^{2}$ with $\beta=10$ being the adjustable hyper-parameter. \cite{xu2018deep} has also used a similar strategy to select a sparse set of differently illuminated images for image-based relighting. However, our focus is on addressing a generalized photometric stereo task. Furthermore, due to the structure of the softmax layer, a larger $\alpha_{r}$ increases the sparsity of each column of $\mathcal{W}$, causing the column to eventually have a single non-zero element at the end corresponding to the optimal chosen sample (see Figure \ref{fig:light_evol}). 

\subsection{Normal Regression Network (NRNet)} The LSNet is trained in conjunction with a Normal Regression Network (NRNet) for each required $M$-light configuration over the Blobby \& Sculpture dataset \cite{chen2018ps}. The feature extraction and aggregation inspire the design of NRNet in several photometric stereo benchmarks such \cite{chen2019self, chen2018ps}. NRNet consists of a shared weight feature extractor consisting of seven convolutional layers that extract features from the images selected by LSNet. The features from each image are then aggregated through a max pool operation (to obtain a global representation shared across images under different lightings) and then passed through four convolutional layers to obtain the surface normal map finally. We selected this architectural design since it reflects the state-of-the-art supervised approaches for addressing photometric stereo. Further, it helps us to avoid any underlying biases due to network architecture, especially when comparing the performance with the existing photometric stereo baselines.

\subsection{Training Details}
LSNet is guided by the normal loss obtained using NRNet during training to learn optimal $M-$light configuration. Both LSNet and NRNet are trained in tandem for around $30$ epochs. However, the model significantly converges at around $10$ epochs. The entire framework is trained to minimize the cosine similarity loss between the estimated ($\widehat{N}_{p}$) and the ground truth (${N}_{p}$) surface normals at every pixel $p$, as described in Equation \ref{eq:5}.
\begin{equation}
    \centering
    \mathcal{L}_{norm}(\widehat{N},N) = \frac{1}{N_{mask}}\sum_{p} \parallel \widehat{N}_{p} - N_{p} \parallel_{2}^{2} 
    \label{eq:5}
\end{equation}
Here, $N_{mask}$ is the sum of pixels in the image mask. We train with the batch size of $32$ with a constant learning rate of $1 \times 10^{-4}$ using Adam optimizer with default parameters on \texttt{NVIDIA RTX 5000} GPU. When trained for $30$ epochs, the model takes $\sim 8$ hours on a single GPU; otherwise, if the user chooses early stopping around $10$ epochs, it takes $\sim 2.5$ to $3$ hours of training time.\footnote{Code will be made available upon acceptance}

\section{Experiments} \label{sec:exp}

\begin{figure}[!ht]
    \centering
    \includegraphics[width=\linewidth]{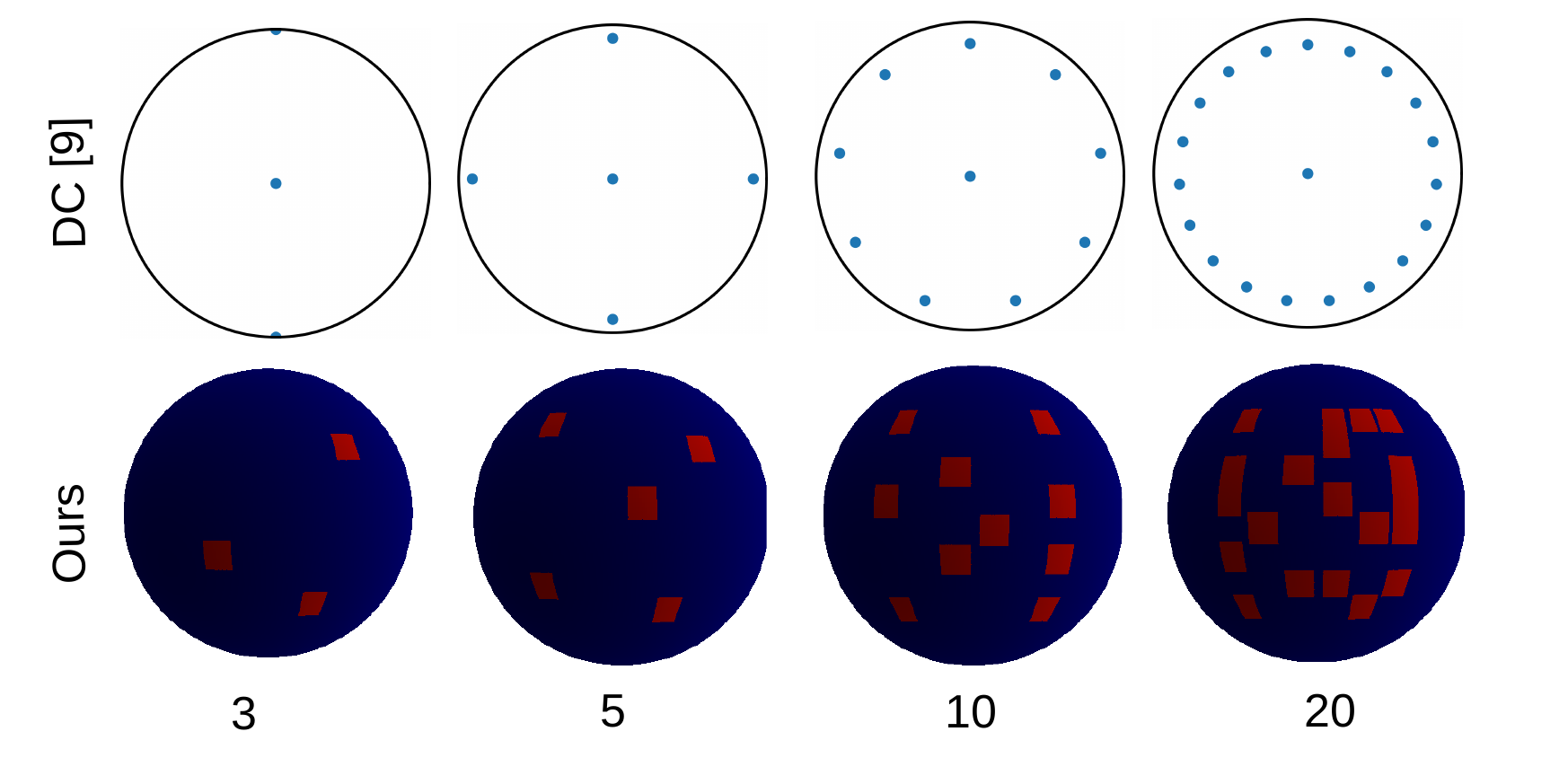}
    \caption{The optimal $M$-light distribution obtained using DC \cite{drbohlav2005optimal} and LIPIDS (ours) for $M = 3,5,10,20$.}
    \label{fig:config}
\end{figure}

While the LSNet and NRNet are trained on the synthetic Blobby \& Sculpture dataset \cite{chen2018ps} for different $M$-light configurations (shown in Figure \ref{fig:config}), we test the learned light configurations with three of the most important photometric stereo backbones PS-FCN \cite{chen2018ps}, CNN-PS \cite{ikehata2018cnn}, and SDM-UniPS \cite{ikehata2023scalable} over four real datasets, namely, DiLiGenT \cite{shi2016benchmark}, DiLiGenT$10^{2}$ \cite{ren2022diligent102} datasets, Gourd\& Apple dataset \cite{alldrin2008photometric}, and Light stage data gallery \cite{debevec2006relighting}. Before inference over any given lighting distribution, we first perform light space discretization by assigning lights to $K=48$ bins as per Algorithm \ref{algo:1}. It is important to note that ours is an offline method that aims to obtain universal optimal lighting through LIPIDS and applies across different photometric stereo backbones at inference.

\begin{table*}[!ht]
\centering

\resizebox{\textwidth}{!}{%
\begin{tabular}{l|l|cccc|cccc|c}
\hline
\multicolumn{1}{c|}{\multirow{2}{*}{\textbf{Datasets}}} & \multicolumn{1}{c|}{\multirow{2}{*}{\textbf{PS Backbone}}} & \multicolumn{4}{c|}{\textbf{10 lights}} & \multicolumn{4}{c|}{\textbf{20 lights}} & \multicolumn{1}{c}{\multirow{2}{*}{\textbf{All Images}}} \\
\multicolumn{1}{c|}{} & \multicolumn{1}{c|}{} & \textbf{DC \cite{drbohlav2005optimal}} & \textbf{TK \cite{tanikawa2022online}} & \textbf{ReLeaPS \cite{chan2023releaps}} & \textbf{Ours} & \textbf{DC \cite{drbohlav2005optimal}} & \textbf{TK \cite{tanikawa2022online}} & \textbf{ReLeaPS \cite{chan2023releaps}} & \textbf{Ours} & \multicolumn{1}{l}{} \\\hline
\multirow{3}{*}{Blobby\cite{johnson2011shape}} & LS \cite{wu2011robust} & 24.63 & 24.42 & 24.72 & \textbf{24.03} & 24.96 & \textbf{23.96} & 24.34 & 24.27 & 24.14 \\
 & PS-FCN \cite{chen2018ps} & 9.46 & 9.1 & 6.94 & \textbf{6.36} & 5.9 & 7.21 & 5.48 & \textbf{5.45} & 5.24 \\  \cline{2-11}
\multicolumn{1}{l}{} & \multicolumn{1}{|l|}{Average} & \multicolumn{1}{c}{17.04} & \multicolumn{1}{c}{16.76} & \multicolumn{1}{c}{15.83} & \multicolumn{1}{c|}{\textbf{15.19}} & \multicolumn{1}{c}{15.43} & \multicolumn{1}{c}{15.585} & \multicolumn{1}{c}{14.91} & \multicolumn{1}{c|}{\textbf{14.86}} & \multicolumn{1}{c}{} \\ \hline
\multirow{3}{*}{Sculpture \cite{chen2018ps}} & LS \cite{wu2011robust} & 26.2 & \textbf{24.99} & 26.07 & 25.53 & 26.38 & \textbf{25.11} & 25.69 & 25.66 & 25.36 \\
 & PS-FCN \cite{chen2018ps} & 9.88 & 11.11 & \textbf{9.59} & 9.77 & 9.23 & 9.56 & 8.54 & \textbf{8.52} & 8.22 \\ \cline{2-11}
\multicolumn{1}{l}{} & \multicolumn{1}{|l|}{Average} & \multicolumn{1}{c}{18.04} & \multicolumn{1}{c}{18.05} & \multicolumn{1}{c}{17.83} & \multicolumn{1}{c|}{\textbf{17.65}} & \multicolumn{1}{c}{17.805} & \multicolumn{1}{c}{17.335} & \multicolumn{1}{c}{17.115} & \multicolumn{1}{c|}{\textbf{17.09}} & \multicolumn{1}{c}{} \\ \hline
\multirow{3}{*}{Bunny \cite{matusik2003data}} & LS \cite{wu2011robust} & 24.71 & 23.38 & 24.45 & 23.85 & 25.43 & 23.79 & 24.68 & 24.71 & 24.33 \\
 & PS-FCN \cite{chen2018ps} & \textbf{6.09} & 8.06 & 8.85 & 6.16 & 5.65 & 6.15 & 8.08 & \textbf{5.4} & 5.25 \\  \cline{2-11}
\multicolumn{1}{l}{} & \multicolumn{1}{|l|}{Average} & \multicolumn{1}{c}{15.4} & \multicolumn{1}{c}{15.72} & \multicolumn{1}{c}{16.65} & \multicolumn{1}{c|}{\textbf{15.01}} & \multicolumn{1}{c}{15.54} & \multicolumn{1}{c}{14.97} & \multicolumn{1}{c}{16.38} & \multicolumn{1}{c|}{\textbf{15.06}} & \multicolumn{1}{c}{} \\ \hline
\multirow{4}{*}{DiLiGenT \cite{shi2016benchmark}} & LS \cite{wu2011robust} & 15.74 & 15.28 & 16.08 & \textbf{15.13} & 15.31 & 15.07 & 15.32 & \textbf{14.78} & 14.67 \\
 & PS-FCN \cite{chen2018ps} & \textbf{8.79} & 8.92 & 8.85 & 8.91 & 8.23 & 8.53 & 8.08 & \textbf{7.82} & 7.76 \\
 & CNN-PS \cite{ikehata2018cnn} & 17.71 & 16.33 & \textbf{12.95} & 13.64 & 13.43 & 14.52 & 11.26 & \textbf{11.23 }& 10.3 \\ \cline{2-11}
\multicolumn{1}{l}{} & \multicolumn{1}{|l|}{Average} & \multicolumn{1}{c}{14.08} & \multicolumn{1}{c}{13.51} & \multicolumn{1}{c}{12.63} & \multicolumn{1}{c|}{\textbf{12.56}} & \multicolumn{1}{c}{12.32} & \multicolumn{1}{c}{12.71} & \multicolumn{1}{c}{11.55} & \multicolumn{1}{c|}{\textbf{11.28}} & \multicolumn{1}{c}{} \\ \hline
\multirow{4}{*}{DiLiGenT$10^{2}$ \cite{ren2022diligent102}} & LS \cite{wu2011robust} & 25.08 & 24.34 & 25.05 & \textbf{24.27} & 25.1 & 23.77 & 24.1 & \textbf{23.72} & 22.82 \\
 & PS-FCN \cite{chen2018ps} & 18.39 & 18.35 & 18.03 & \textbf{18.01} & \textbf{18.32} & 18.42 & 18.5 & 18.68 & 18.31 \\
 & CNN-PS \cite{ikehata2018cnn} & 27.81 & 20.8 & 22.2 & \textbf{19.78} & 22.01 & \textbf{19.33} & 19.7 & 19.38 & 19.24 \\ \cline{2-11}
\multicolumn{1}{l}{} & \multicolumn{1}{|l|}{Average} & \multicolumn{1}{c}{23.76} & \multicolumn{1}{c}{21.16} & \multicolumn{1}{c}{21.76} & \multicolumn{1}{c|}{\textbf{20.69}} & \multicolumn{1}{c}{21.81} & \multicolumn{1}{c}{20.51} & \multicolumn{1}{c}{20.77} & \multicolumn{1}{c|}{\textbf{20.59}} & \multicolumn{1}{c}{} \\ \hline
\end{tabular}%
}
\caption{Quantitative comparison of different illumination planning methods across different datasets and different photometric stereo backbones for $10$ and $20$-light configuration. The values represent the mean angular error (MAE).}
\label{tab:1}
\end{table*}

\begin{figure*}[!ht]
\centering

\begin{tabular}{cccc}
    \includegraphics[width=.45\linewidth]{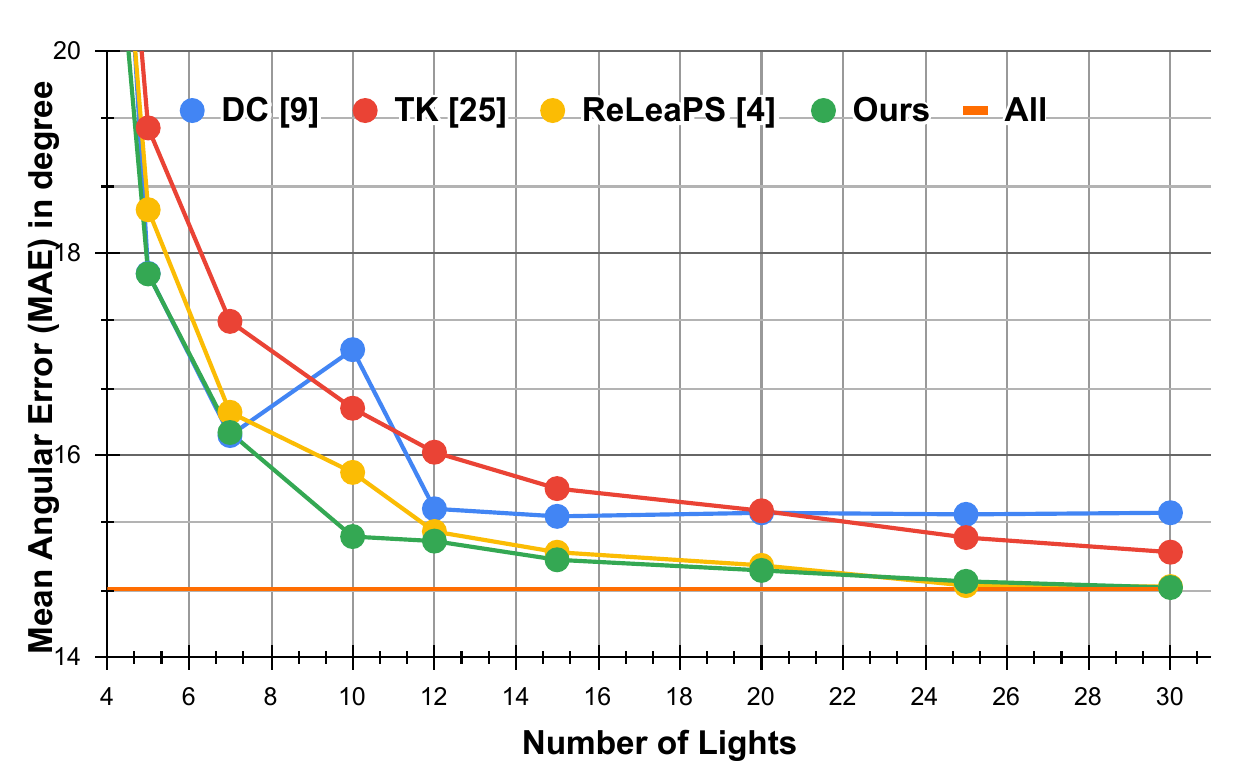}      &    \includegraphics[width=.45\linewidth]{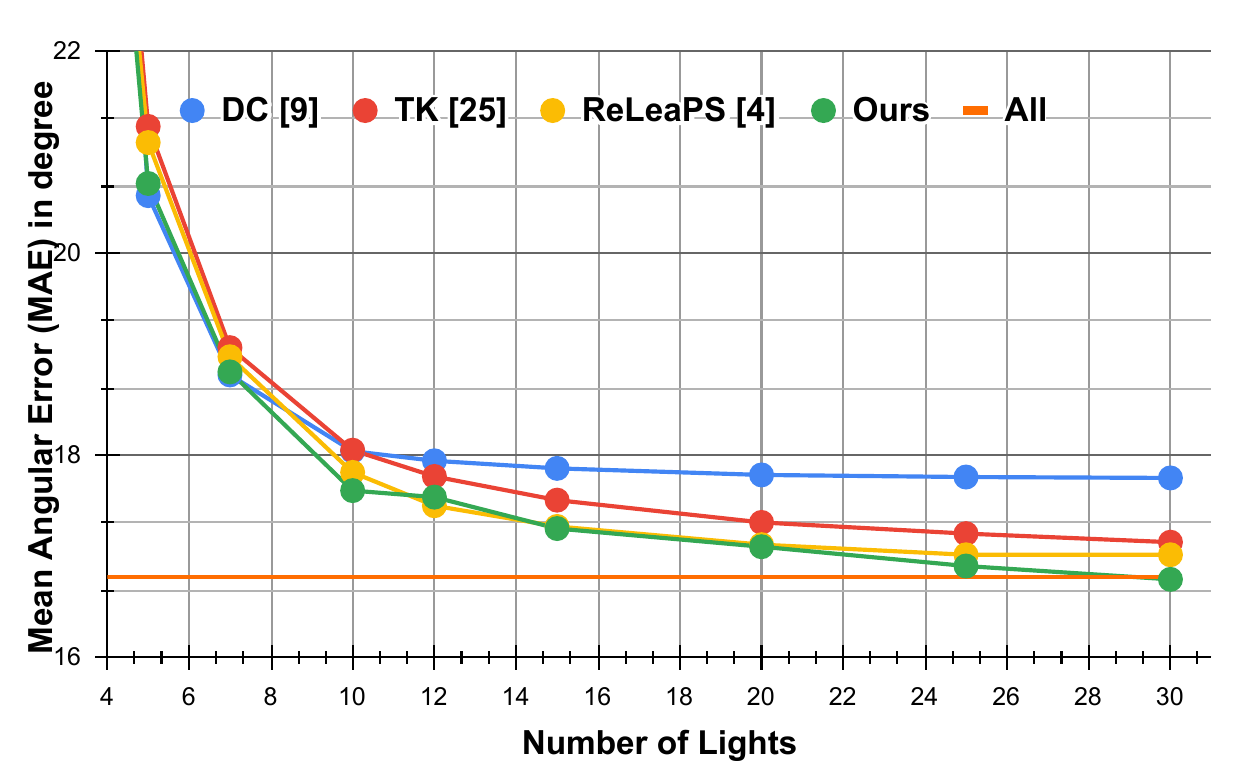}      & \\   
    (a) Blobby  & (b) Sculpture  & \\
    \includegraphics[width=.45\linewidth]{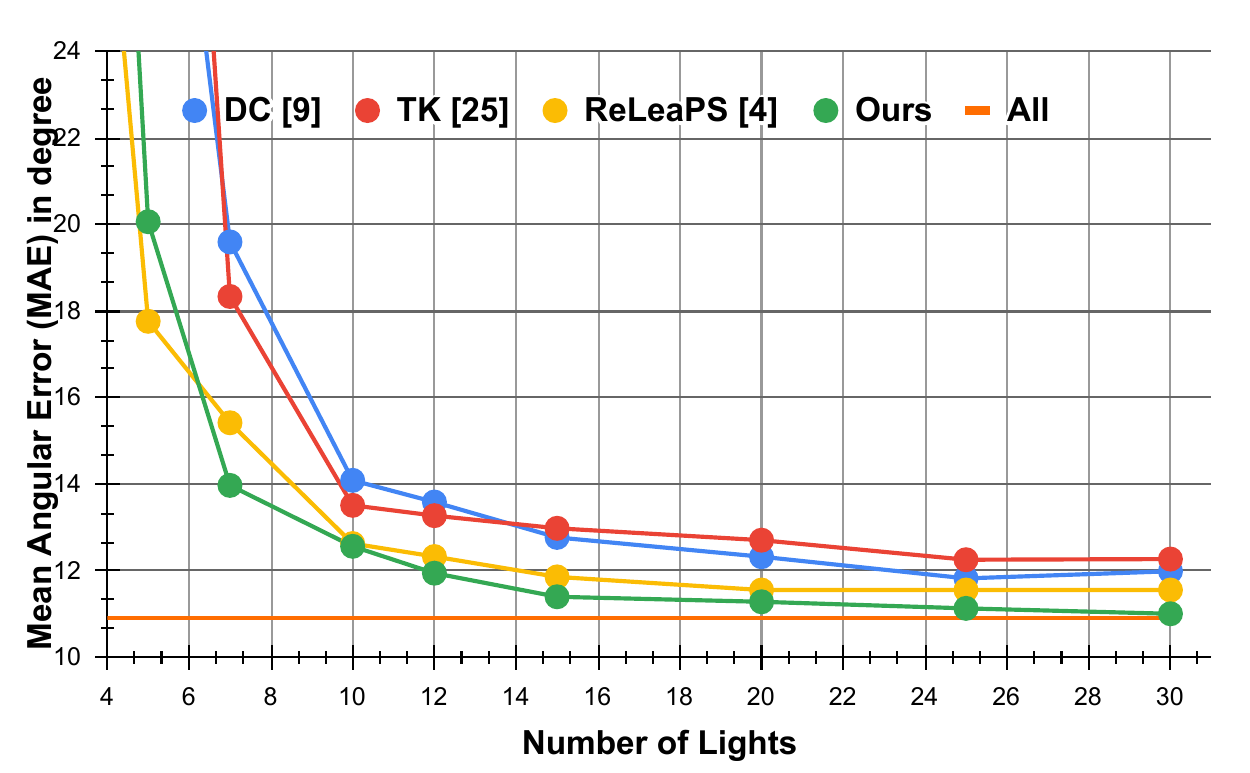} &
    \includegraphics[width=.45\linewidth]{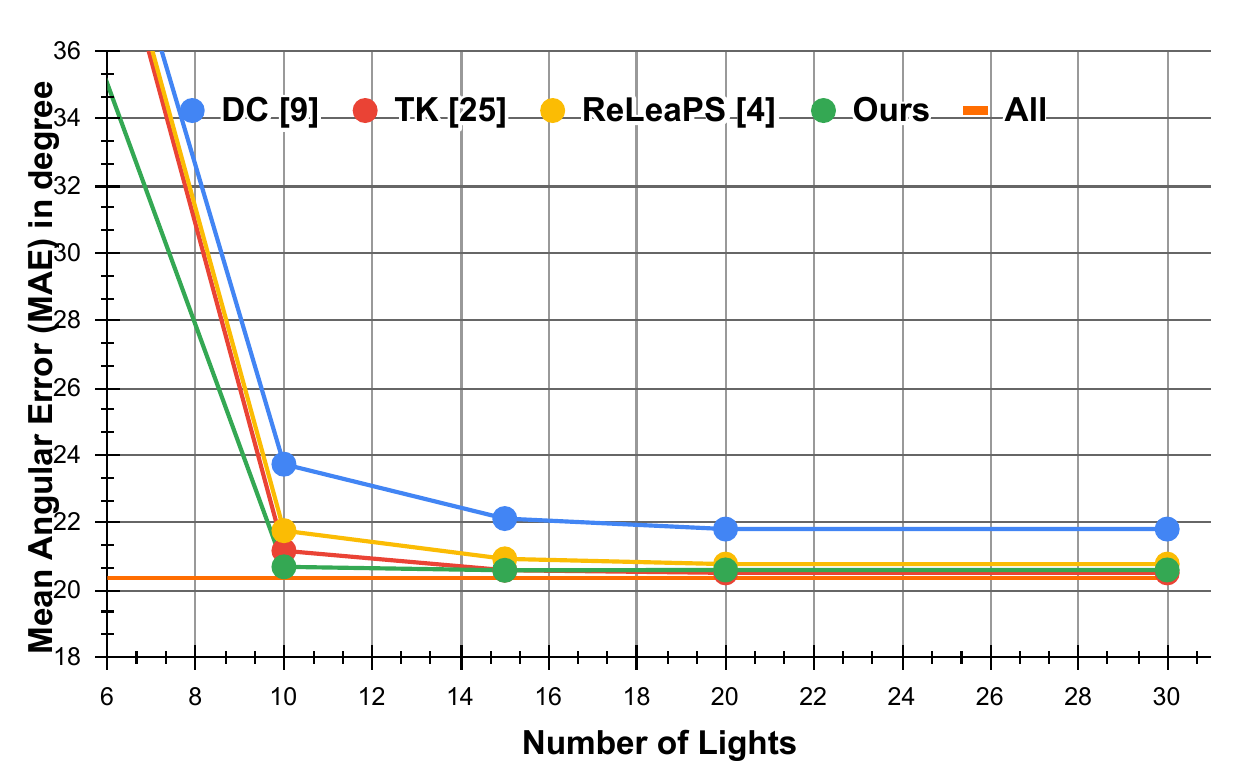}    \\
     (a) DiLiGenT & (b) DiLiGenT$10^{2}$ \\
   
\end{tabular}
\caption{Quantitative evaluation of illumination planning methods with the different number of light directions on (a,b) synthetic: Blobby and Sculpture \cite{johnson2011shape} and (c,d) real
benchmark datasets: DiLiGenT \cite{shi2016benchmark} and DiLiGenT$10^{2}$ \cite{ren2022diligent102}. We report the mean angular error averaged over LS \cite{wu2011robust}, PS-FCN \cite{chen2018ps}, and CNN-PS \cite{ikehata2018cnn} backbones for real data. However, for synthetic data, the average is taken over LS \cite{wu2011robust} and PS-FCN \cite{chen2018ps}.}
\label{fig:plots}
\end{figure*}


   


\subsection{Quantitative Evaluation}
Table \ref{tab:1} compares the performance of our proposed method with three other illumination planning methods — DC \cite{drbohlav2005optimal}, TK \cite{tanikawa2022online}, and ReLeaPS \cite{chan2023releaps} — using $10$ and $20$ lights on both synthetic and real datasets. For the Blobby \& Sculpture dataset \cite{johnson2011shape}, testing was conducted on a held-out test set. We selected the observation configuration with vertical light direction to compare our method with the best-performing setting of DC \cite{drbohlav2005optimal}. Additionally, we compared the performance of our learned lighting configuration with other photometric stereo backbones. Our offline method outperformed existing methods and closely matched the performance of the online method ReLeaPS. On the DiLiGent$10^{2}$ dataset, our method clearly outperformed other illumination planning methods with a 10-light configuration but slightly lagged over CNN-PS \cite{ikehata2018cnn} and PS-FCN \cite{chen2018ps} with a 20-light configuration. The opposite trend was observed with the DiLiGenT dataset, likely due to different training paradigms, per-pixel vs. all-pixel in CNN-PS and PS-FCN, respectively. Overall, our method showed a strong lead when averaged across both methods and datasets. While ReLeaPS under-performs due to narrower light distribution on the DiLiGenT dataset \cite{chan2023releaps}, our method is not constrained by such lighting distribution limitations. Averaging performance across both 10-light and 20-light settings, our technique consistently performed best across all datasets and backbones, indicating its broader applicability. Furthermore, we compare the performance of all the methods over the different number of light directions. Figure \ref{fig:plots} shows the comparison across both synthetic and real datasets. LIPIDS consistently outperforms others, including ReLeaPS, for any $K > 5$. The performance on the DiLiGenT and DiLiGenT$10^{2}$ begins to saturate beyond $20$ lights, indicating that a maximum of $20$ lights could be enough to understand the geometry well. 

\begin{figure*}[!ht]
    \centering
    \includegraphics[width=\linewidth]{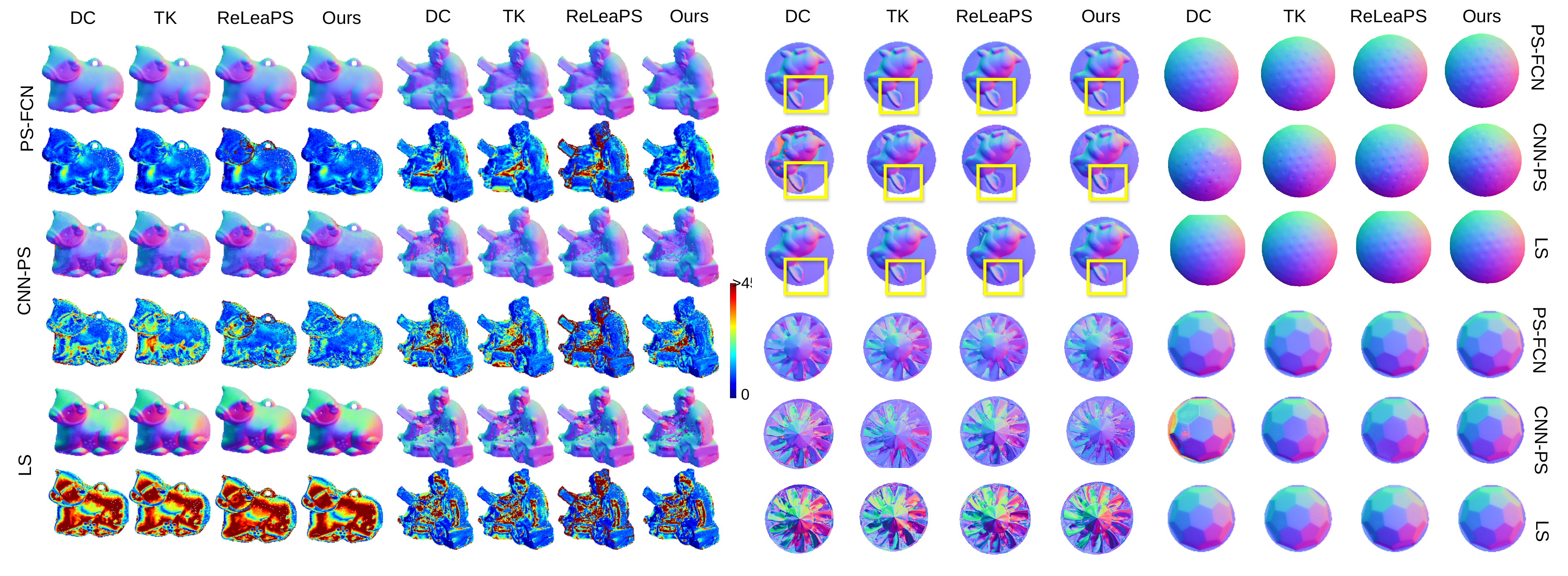}
    \caption{Qualitative comparison of different illumination planning methods over objects from the DiLiGenT dataset \cite{shi2016benchmark} (first-half) and DiLiGenT$10^{2}$ \cite{ren2022diligent102} dataset (second-half).}
    \label{fig:qual_res_1}
\end{figure*}

\textbf{Comparison with other sampling strategies.} To evaluate the quality of our learned lighting configuration, we compare it with two heuristic-based strategies —random dart throwing and k-means clustering the light direction of the Blobby \& Sculpture test set with a PS-FCN backbone. Random choice of 10 lights resulted in a high MAE of $10.07^{\circ}$. k-means clustering ($k=10$) performs slightly better with an MAE of $9.53^{\circ}$, surpassing DC \cite{drbohlav2005optimal} ($9.67^{\circ}$) and TK \cite{tanikawa2022online} ($10.11^{\circ}$), but still falling behind ReLeaPS \cite{chan2023releaps} and LIPIDS, which achieve MAEs of $8.26^{\circ}$ and $8.06^{\circ}$, respectively. To ensure that random choice is well-distributed, we specify a minimal threshold on the distance between two samples and also from the boundary of the light domain to avoid accumulation near the boundary of the light domain. Moreover, we found that k-means clustering converges to a distribution biased by the density of the light sampling. Figure \ref{fig:config} shows that the different learned $M$-light configurations through LIPIDS are less biased compared to the heuristic-based method DC \cite{drbohlav2005optimal}. 

\textbf{Comparison with ReLeaPS.} Our performance has been nearly equal to or closely surpassing that of the ReLeaPS. While ReLeaPS, as an online method, benefits from obtaining optimal lighting per object tailored to specific object properties like reflectance and material, it often demands extensive convergence time and a complex setup. ReLeaPS requires approximately $1.5$ days for training and around $20$ seconds per object during inference (evaluated on the same hardware, \texttt{NVIDIA RTX A5000}). In contrast, our offline method requires only about $3$ hours for training and roughly a second for inference.  Therefore, we believe LIPIDS could be a faster alternative to online methods.

\textbf{Improving PS backbones.} Recently, SDM-UniPS \cite{ikehata2023scalable} demonstrated the need for properly varying illumination across different images to understand the surface normals better. Since it can estimate surface normal with any number of images, we evaluate SDM-UniPS with LSNet and compare the resulting MAE with SDM-UniPS evaluated for $5$, $10$, and $20$ lights. We observed that the learned lighting configuration through LSNet helped enhance the normal estimation (see Table \ref{tab:sdm}). This indicates that LSNet (or LIPIDS in general) can be integrated with an existing photometric stereo network, even while training, to utilize the images in the dataset optimally.

While capturing more images under different lighting conditions is always an option, the main goal is to determine the optimal placement of these light sources for the best results. LIPIDS not only provides the optimal configuration for a given number of lights but also identifies a reasonably minimal number of lights applicable across various datasets.


\begin{table}[!ht]
\centering

\resizebox{\linewidth}{!}{%
\begin{tabular}{l|c|c|c|c}\hline
Datasets & \multicolumn{1}{c|}{Method} & 5 lights & 10 lights & 20 lights  \\\hline
\multirow{2}{*}{DiLiGenT \cite{shi2016benchmark}} 
  & SDM-UniPS \cite{ikehata2023scalable} & 7.51 & 6.97 & 6.51 \\
  & LSNet + SDM-UniPS \cite{ikehata2023scalable} & \textbf{7.02} & \textbf{6.75} & \textbf{6.34} \\\hline 

\end{tabular}%
}
\caption{Quantitative evaluation of combining LSNet with SDM-UniPS \cite{ikehata2023scalable}.}
\label{tab:sdm}
\end{table}



\subsection{Qualitative Evaluation}
We now evaluate the effectiveness of the LIPIDS framework primarily across the real-world datasets in Figure \ref{fig:qual_res_1}. Shadows play a pivotal role in deciding the orientation of surface normals. The \textsf{READING} object from the DiLiGenT \cite{shi2016benchmark} dataset exhibits global illumination effects prominently at the object's centre. Our method better resolves the surface normals in such regions affected by shadows and inter-reflections (see Figure \ref{fig:qual_res_1}, left) across all the backbones. We also compare the performance over the DiLiGenT$10^{2}$ \cite{ren2022diligent102} dataset in Figure \ref{fig:qual_res_1} (right). While the qualitative effects are not prominent in smooth and simple objects, we observe clear improvements in relatively complex objects (marked in YELLOW coloured boxes). Further, we observe similar performance over the Light Stage Data Gallery \cite{debevec2006relighting} and Gourd \& Apple dataset \cite{alldrin2008photometric} across different methods for which the results are shown in the supplementary material. Furthermore, we also analyze the effect of varying numbers of lights over different methods and real-world datasets in the supplementary material.



\section{Conclusion} \label{sec:conclusion}
We introduced an effective discretized light-space sampling technique to address illumination planning for generalized photometric stereo in an offline manner. We demonstrated that the learned light configuration is more optimal (and minimal) than other illumination planning methods when evaluated with different photometric stereo backbones and applied over different real-world datasets with arbitrary lighting distribution. We show that it can also improve the model performance through the learned lighting configuration, especially under sparse settings, upon integrating with an existing photometric stereo backbone. One could train LSNet along with any existing backbones and optimally utilize the existing datasets for better performance instead of relying on large-scale sophisticated data capture. We hope the simplicity and applicability of the proposed framework will draw more attention to optimally minifying photometric stereo through illumination planning.

\section*{Acknowledgements}
We would like to acknowledge the generous support of the Prime Minister Research Fellowship (PMRF) grant and the Jibaben Patel Chair in Artificial Intelligence for the completion of this work.

{\small
\bibliographystyle{ieee_fullname}
\bibliography{egbib}

\begin{thebibliography}{10}\itemsep=-1pt

\bibitem{ackermann2012photometric}
Jens Ackermann, Fabian Langguth, Simon Fuhrmann, and Michael Goesele.
\newblock Photometric stereo for outdoor webcams.
\newblock In {\em 2012 IEEE conference on computer vision and pattern recognition}, pages 262--269. IEEE, 2012.

\bibitem{alldrin2008photometric}
Neil Alldrin, Todd Zickler, and David Kriegman.
\newblock Photometric stereo with non-parametric and spatially-varying reflectance.
\newblock In {\em 2008 IEEE Conference on Computer Vision and Pattern Recognition}, pages 1--8. IEEE, 2008.

\bibitem{chakrabarti2016learning}
Ayan Chakrabarti.
\newblock Learning sensor multiplexing design through back-propagation.
\newblock {\em Advances in Neural Information Processing Systems}, 29, 2016.

\bibitem{chan2023releaps}
Jun~Hoong Chan, Bohan Yu, Heng Guo, Jieji Ren, Zongqing Lu, and Boxin Shi.
\newblock Releaps: Reinforcement learning-based illumination planning for generalized photometric stereo.
\newblock In {\em Proceedings of the IEEE/CVF International Conference on Computer Vision}, pages 9167--9175, 2023.

\bibitem{chen2019self}
Guanying Chen, Kai Han, Boxin Shi, Yasuyuki Matsushita, and Kwan-Yee~K Wong.
\newblock Self-calibrating deep photometric stereo networks.
\newblock In {\em Proceedings of the IEEE Conference on Computer Vision and Pattern Recognition}, pages 8739--8747, 2019.

\bibitem{chen2018ps}
Guanying Chen, Kai Han, and Kwan-Yee~K Wong.
\newblock Ps-fcn: A flexible learning framework for photometric stereo.
\newblock In {\em Proceedings of the European Conference on Computer Vision (ECCV)}, pages 3--18, 2018.

\bibitem{debevec2006relighting}
Paul Debevec.
\newblock Relighting human locomotion.
\newblock In {\em ACM SIGGRAPH 2006 Computer animation festival}, pages 263--es. 2006.

\bibitem{debevec2000acquiring}
Paul Debevec, Tim Hawkins, Chris Tchou, Haarm-Pieter Duiker, Westley Sarokin, and Mark Sagar.
\newblock Acquiring the reflectance field of a human face.
\newblock In {\em Proceedings of the 27th annual conference on Computer graphics and interactive techniques}, pages 145--156, 2000.

\bibitem{drbohlav2005optimal}
Ondrej Drbohlav and Mike Chantler.
\newblock On optimal light configurations in photometric stereo.
\newblock In {\em Tenth IEEE International Conference on Computer Vision (ICCV'05) Volume 1}, volume~2, pages 1707--1712. IEEE, 2005.

\bibitem{enomoto2020photometric}
Kenji Enomoto, Michael Waechter, Kiriakos~N Kutulakos, and Yasuyuki Matsushita.
\newblock Photometric stereo via discrete hypothesis-and-test search.
\newblock In {\em Proceedings of the IEEE/CVF Conference on Computer Vision and Pattern Recognition}, pages 2311--2319, 2020.

\bibitem{gardi2022optimal}
Hamza Gardi, Sebastian~F Walter, and Christoph~S Garbe.
\newblock An optimal experimental design approach for light configurations in photometric stereo.
\newblock {\em arXiv preprint arXiv:2204.05218}, 2022.

\bibitem{Guo_Ren_Wang_2024_CVPR}
Heng Guo, Jieji Ren, Feishi Wang, Mingjun Ren, Boxin Shi, and Matsushita Yasuyuki.
\newblock Diligenrt: A photometric stereo dataset with quantifed roughness and translucency.
\newblock In {\em Proceedings of the IEEE/CVF Computer Vision and Pattern Recongnition (CVPR)}, pages xxxxx--xxxxx, June 2024.

\bibitem{ikehata2018cnn}
Satoshi Ikehata.
\newblock Cnn-ps: Cnn-based photometric stereo for general non-convex surfaces.
\newblock In {\em Proceedings of the European conference on computer vision (ECCV)}, pages 3--18, 2018.

\bibitem{ikehata2022ps}
Satoshi Ikehata.
\newblock Ps-transformer: Learning sparse photometric stereo network using self-attention mechanism.
\newblock {\em arXiv preprint arXiv:2211.11386}, 2022.

\bibitem{ikehata2023scalable}
Satoshi Ikehata.
\newblock Scalable, detailed and mask-free universal photometric stereo.
\newblock In {\em Proceedings of the IEEE/CVF Conference on Computer Vision and Pattern Recognition}, pages 13198--13207, 2023.

\bibitem{iwaguchi2023surface}
Takafumi Iwaguchi and Hiroshi Kawasaki.
\newblock Surface normal estimation from optimized and distributed light sources using dnn-based photometric stereo.
\newblock In {\em Proceedings of the IEEE/CVF Winter Conference on Applications of Computer Vision}, pages 311--320, 2023.

\bibitem{johnson2011shape}
Micah~K Johnson and Edward~H Adelson.
\newblock Shape estimation in natural illumination.
\newblock In {\em CVPR 2011}, pages 2553--2560. IEEE, 2011.

\bibitem{ju2024deep}
Yakun Ju, Kin-Man Lam, Wuyuan Xie, Huiyu Zhou, Junyu Dong, and Boxin Shi.
\newblock Deep learning methods for calibrated photometric stereo and beyond.
\newblock {\em IEEE Transactions on Pattern Analysis and Machine Intelligence}, 2024.

\bibitem{li2019learning}
Junxuan Li, Antonio Robles-Kelly, Shaodi You, and Yasuyuki Matsushita.
\newblock Learning to minify photometric stereo.
\newblock In {\em Proceedings of the IEEE/CVF Conference on Computer Vision and Pattern Recognition}, pages 7568--7576, 2019.

\bibitem{lichy2021shape}
Daniel Lichy, Jiaye Wu, Soumyadip Sengupta, and David~W Jacobs.
\newblock Shape and material capture at home.
\newblock In {\em Proceedings of the IEEE/CVF Conference on Computer Vision and Pattern Recognition}, pages 6123--6133, 2021.

\bibitem{logothetis2021px}
Fotios Logothetis, Ignas Budvytis, Roberto Mecca, and Roberto Cipolla.
\newblock Px-net: Simple and efficient pixel-wise training of photometric stereo networks.
\newblock In {\em Proceedings of the IEEE/CVF International Conference on Computer Vision}, pages 12757--12766, 2021.

\bibitem{matusik2003data}
Wojciech Matusik.
\newblock {\em A data-driven reflectance model}.
\newblock PhD thesis, Massachusetts Institute of Technology, 2003.

\bibitem{ren2022diligent102}
Jieji Ren, Feishi Wang, Jiahao Zhang, Qian Zheng, Mingjun Ren, and Boxin Shi.
\newblock Diligent102: A photometric stereo benchmark dataset with controlled shape and material variation.
\newblock In {\em Proceedings of the IEEE/CVF Conference on Computer Vision and Pattern Recognition}, pages 12581--12590, 2022.

\bibitem{santo2017deep}
Hiroaki Santo, Masaki Samejima, Yusuke Sugano, Boxin Shi, and Yasuyuki Matsushita.
\newblock Deep photometric stereo network.
\newblock In {\em Proceedings of the IEEE international conference on computer vision workshops}, pages 501--509, 2017.

\bibitem{shi2016benchmark}
Boxin Shi, Zhe Wu, Zhipeng Mo, Dinglong Duan, Sai-Kit Yeung, and Ping Tan.
\newblock A benchmark dataset and evaluation for non-lambertian and uncalibrated photometric stereo.
\newblock In {\em Proceedings of the IEEE Conference on Computer Vision and Pattern Recognition}, pages 3707--3716, 2016.

\bibitem{tanikawa2022online}
Hirochika Tanikawa, Ryo Kawahara, and Takahiro Okabe.
\newblock Online illumination planning for shadow-robust photometric stereo.
\newblock In {\em International Workshop on Frontiers of Computer Vision}, pages 80--93. Springer, 2022.

\bibitem{tiwari2022deepps2}
Ashish Tiwari and Shanmuganathan Raman.
\newblock Deepps2: Revisiting photometric stereo using two differently illuminated images.
\newblock In {\em European Conference on Computer Vision}, pages 129--145. Springer, 2022.

\bibitem{tiwari2022lerps}
Ashish Tiwari and Shanmuganathan Raman.
\newblock Lerps: Lighting estimation and relighting for photometric stereo.
\newblock In {\em ICASSP 2022-2022 IEEE International Conference on Acoustics, Speech and Signal Processing (ICASSP)}, pages 2060--2064. IEEE, 2022.

\bibitem{wang2023diligent}
Feishi Wang, Jieji Ren, Heng Guo, Mingjun Ren, and Boxin Shi.
\newblock Diligent-pi: Photometric stereo for planar surfaces with rich details-benchmark dataset and beyond.
\newblock In {\em Proceedings of the IEEE/CVF International Conference on Computer Vision}, pages 9477--9487, 2023.

\bibitem{woodham1980photometric}
Robert~J Woodham.
\newblock Photometric method for determining surface orientation from multiple images.
\newblock {\em Optical engineering}, 19(1):139--144, 1980.

\bibitem{wu2011robust}
Lun Wu, Arvind Ganesh, Boxin Shi, Yasuyuki Matsushita, Yongtian Wang, and Yi Ma.
\newblock Robust photometric stereo via low-rank matrix completion and recovery.
\newblock In {\em Computer Vision--ACCV 2010: 10th Asian Conference on Computer Vision, Queenstown, New Zealand, November 8-12, 2010, Revised Selected Papers, Part III 10}, pages 703--717. Springer, 2011.

\bibitem{xu2018deep}
Zexiang Xu, Kalyan Sunkavalli, Sunil Hadap, and Ravi Ramamoorthi.
\newblock Deep image-based relighting from optimal sparse samples.
\newblock {\em ACM Transactions on Graphics (ToG)}, 37(4):1--13, 2018.

\bibitem{yao2020gps}
Zhuokun Yao, Kun Li, Ying Fu, Haofeng Hu, and Boxin Shi.
\newblock Gps-net: Graph-based photometric stereo network.
\newblock {\em Advances in Neural Information Processing Systems}, 33:10306--10316, 2020.

\bibitem{zheng2019spline}
Qian Zheng, Yiming Jia, Boxin Shi, Xudong Jiang, Ling-Yu Duan, and Alex~C Kot.
\newblock Spline-net: Sparse photometric stereo through lighting interpolation and normal estimation networks.
\newblock In {\em Proceedings of the IEEE/CVF International Conference on Computer Vision}, pages 8549--8558, 2019.

\end{thebibliography}
}

\newpage
\section*{Supplementary: Additional Results}
\begin{figure}[h]
    \centering
    \includegraphics[width=\linewidth]{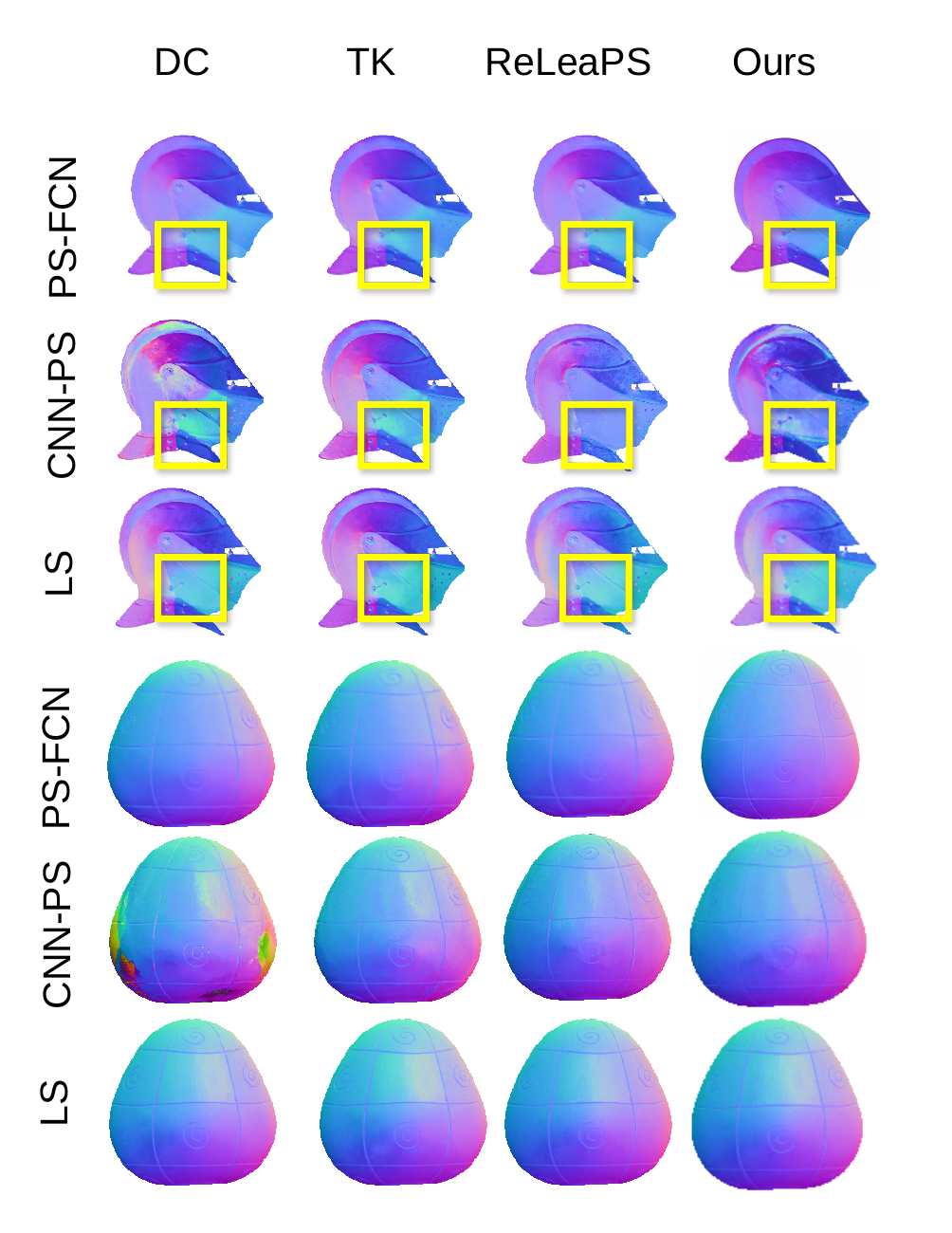}
    \caption{Qualitative comparison of variation of number of lights on the performance of different frameworks and different illumination planning methods over objects from  Light Stage Data Gallery \cite{debevec2006relighting} and Gourd \& Apple \cite{alldrin2008photometric} datasets.}
    \label{fig:qual_res_1}
    
\end{figure}

\begin{figure*}[!ht]
    \centering
    \includegraphics[width=\linewidth]{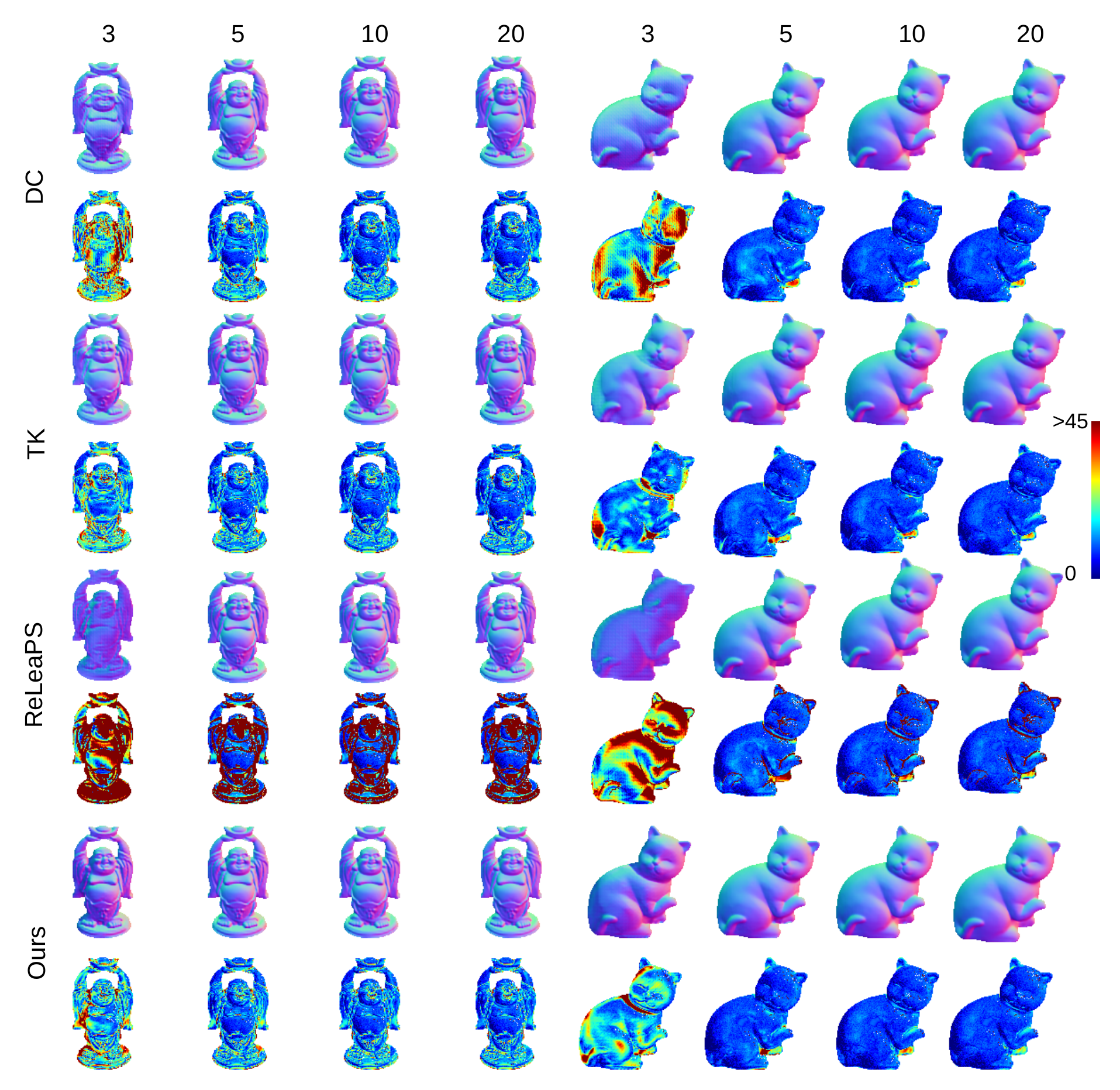}
    \caption{Qualitative comparison of variation of number of lights on the performance of different frameworks and different illumination planning methods over objects from the DiLiGenT dataset \cite{shi2016benchmark} }
    \label{fig:qual_res_4}
\end{figure*}
\begin{figure*}[h]
    \centering
    \includegraphics[width=\linewidth]{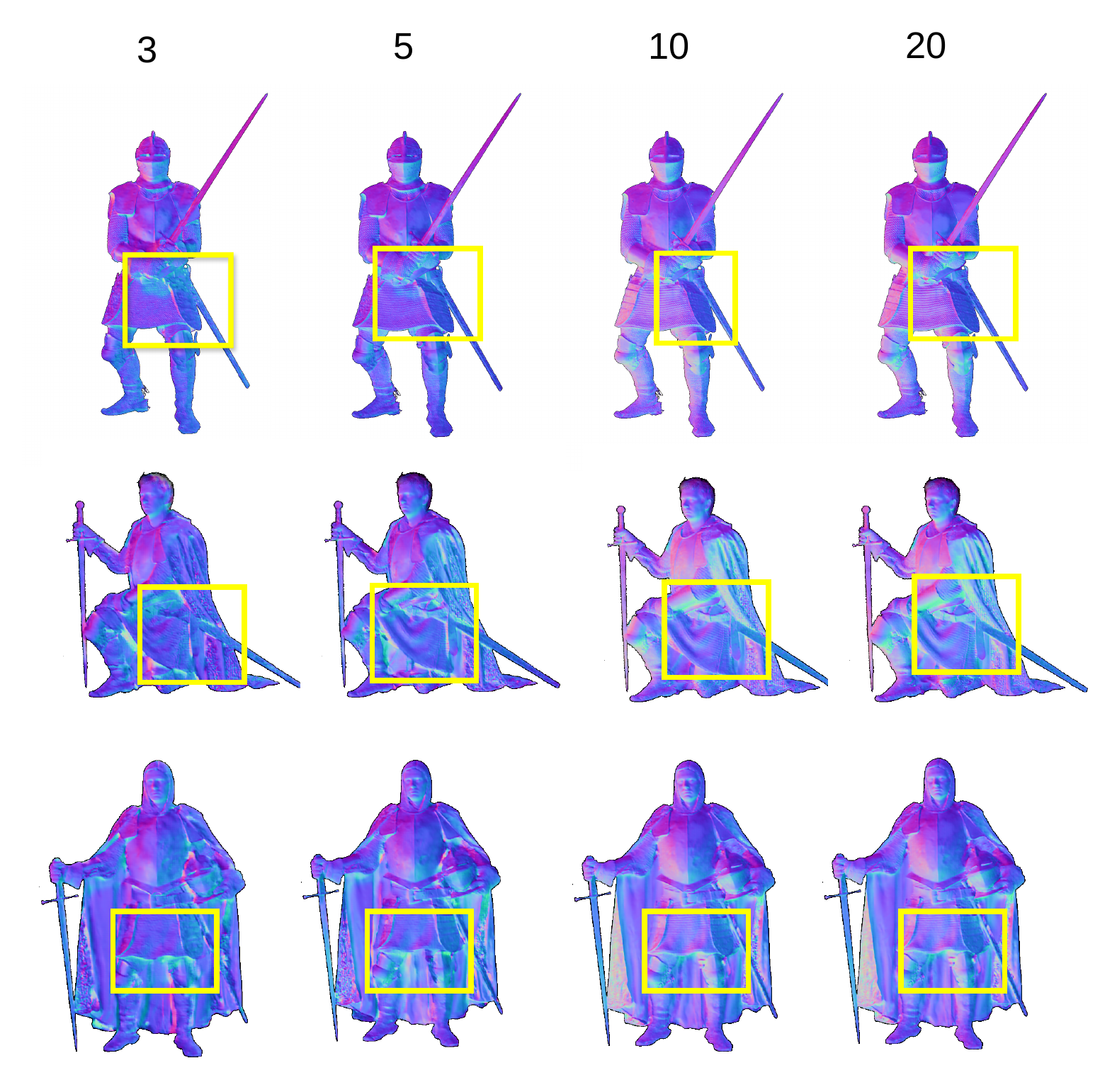}
    \caption{Qualitative comparison of variation of number of lights on the performance of LSNet + PS-FCN framework over objects from the Light Stage Data Gallery \cite{debevec2006relighting} dataset.}
    \label{fig:qual_res_2}
\end{figure*}
\begin{figure*}[h]
    \centering
    \includegraphics[width=\linewidth]{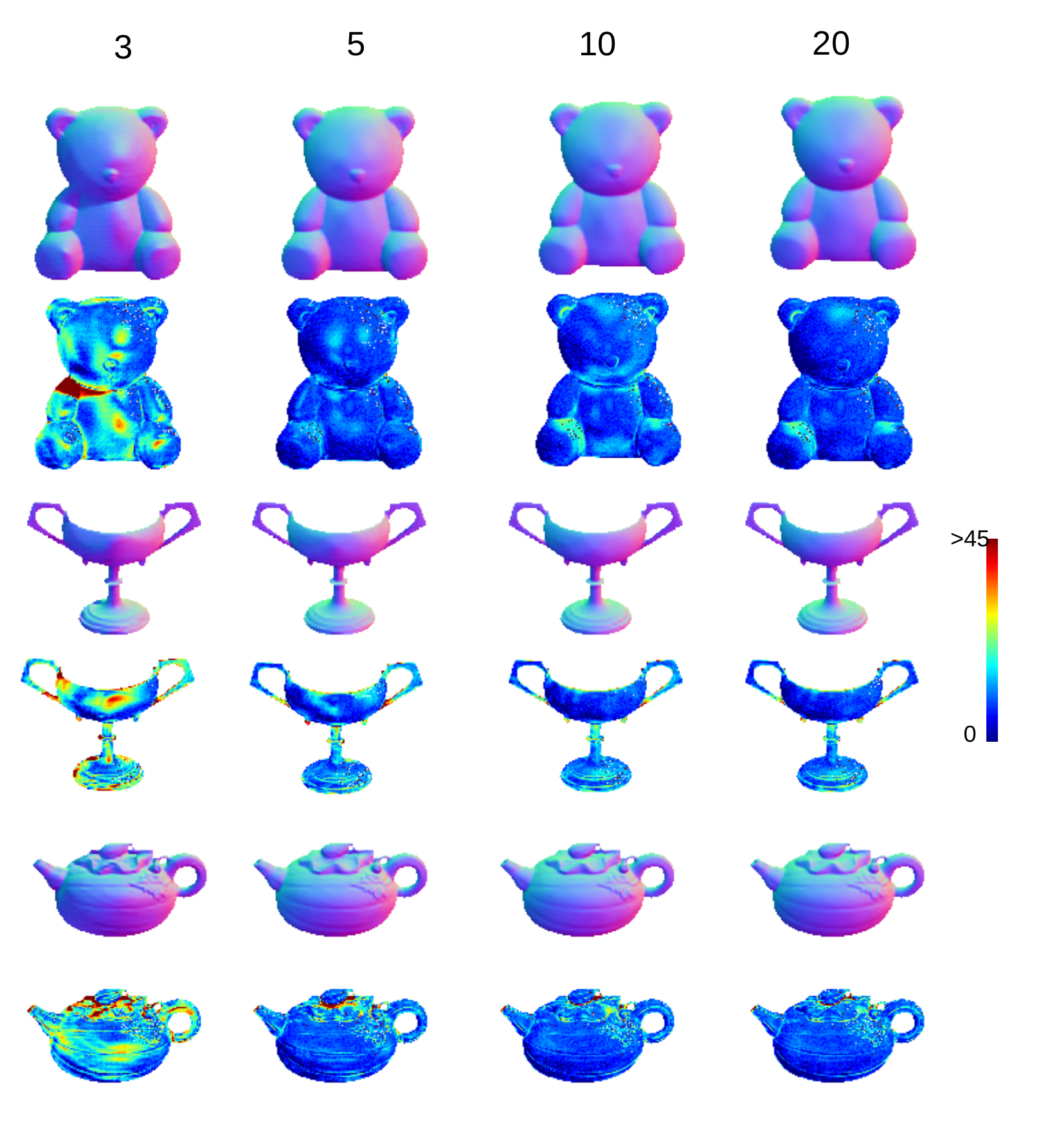}
    \caption{Qualitative comparison of variation of number of lights on the performance of LSNet + PS-FCN framework over objects from the DiLiGenT \cite{shi2016benchmark} dataset.}
    \label{fig:qual_res_3}
\end{figure*}


\end{document}